\documentclass[letterpaper, 10 pt, conference]{ieeeconf}  % Comment this line out if you need a4paper

\IEEEoverridecommandlockouts                              % This command is only needed if 
\usepackage{amsmath, amssymb}

\usepackage{booktabs} 
\usepackage{array}   
\usepackage{graphicx} 
\usepackage{caption}  
\usepackage{makecell}
\usepackage{multirow}
\usepackage{adjustbox}
\usepackage{hyperref}
\title{\LARGE \bf
ActMVS: Active Scene Reconstruction with Monocular Multi-View Stereo
}
\author{
  Guo~Pu$^{1*}$,
  \thanks{This work was supported by National Natural Science Foundation of China (Grant No.: 62372015),
  Leading Projects in Key Research Fields of Language Funded by the National Language Commission,
  and Key Laboratory of Intelligent Press Media Technology.}%
  \thanks{$^{1}$Wangxuan Institute of Computer Technology, Peking University}
  \thanks{*Equal contribution.}%
  Yixuan~Han$^{1*}$,
  Zhouhui~Lian$^{1\dagger}$
  \thanks{$^{\dagger}$Corresponding author: \tt\small lianzhouhui@pku.edu.cn}
}

\begin{document}

\maketitle
\thispagestyle{empty}
\pagestyle{empty}

%%%%%%%%%%%%%%%%%%%%%%%%%%%%%%%%%%%%%%%%%%%%%%%%%%%%%%%%%%%%%%%%%%%%%%%%%%%%%%%%
\begin{abstract}

Active scene reconstruction enables robots/UAVs to autonomously plan trajectories and reconstruct environments without costly manual data acquisition. Unlike passive methods, active reconstruction requires real-time construction of high-confidence occupancy maps for collision-free navigation. Existing approaches rely on depth sensors for occupancy map updates, increasing platform cost and weight. To advance spatial intelligence, we aim for a vision-only monocular  solution. However, current monocular scene reconstruction methods operate offline and fail to deliver globally consistent dense depth at the frame rates required for robots/UAVs navigation. To bridge this gap, we introduce ActMVS, the first framework for monocular active reconstruction. Our framework integrates a view factor graph construction for informed Multi-View Stereo depth prediction, along with a global depth optimization, to enable the online generation of high-quality, globally consistent dense depth maps. This enables monocular robots/UAVs to maintain reliable occupancy maps for safe trajectory planning during reconstruction. Experiments on Replica datasets demonstrate performance competitive with RGB-D methods. Our code and data are available at \url{https://github.com/TrickyGo/ActMVS}.

\end{abstract}

%%%%%%%%%%%%%%%%%%%%%%%%%%%%%%%%%%%%%%%%%%%%%%%%%%%%%%%%%%%%%%%%%%%%%%%%%%%%%%%%
\section{Introduction}
\label{sec:introduction}

\begin{figure*}[t]
  \centering
  \includegraphics[width=0.85\textwidth]{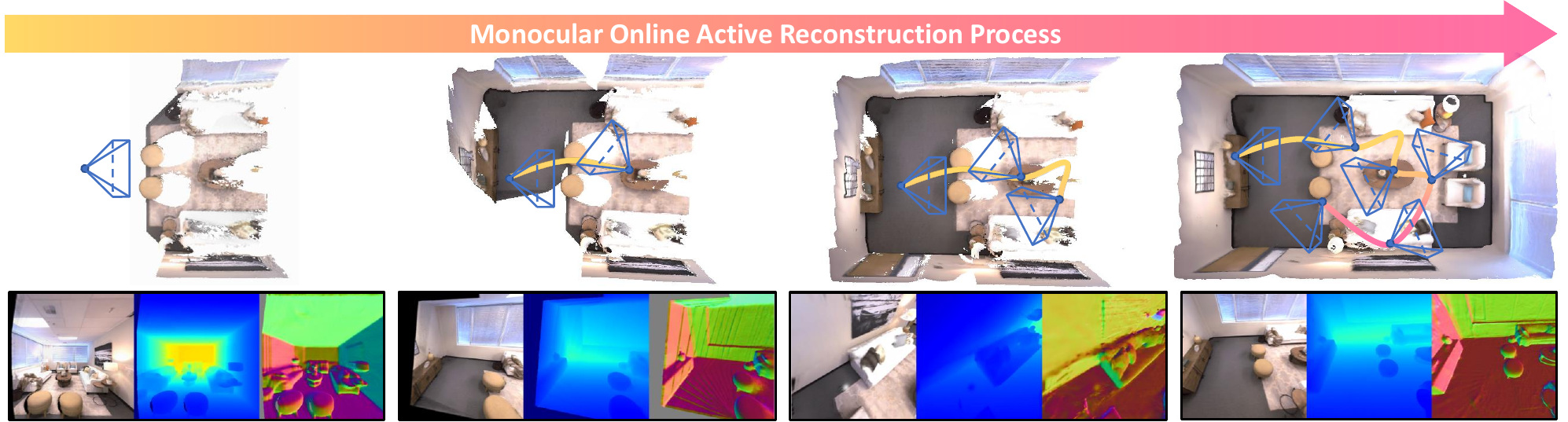}
  % \vspace{-8pt}
  \caption{The ActMVS monocular active reconstruction process which shows the intermediate mesh, camera trajectory, and rendered image with depth and surface normal maps.}
  \vspace{-15pt}
  \label{fig:teaser}
\end{figure*}

Active scene reconstruction is a critical capability for robots/UAVs (Unmanned Aerial Vehicles) to achieve autonomous navigation and reconstruction in unknown environments~\cite{chen2011active}, indispensable for applications like industrial inspection and disaster response, where manual data acquisition is prohibitively costly. 
While passive reconstruction methods such as SfM~\cite{schonberger2016structure} and SLAM~\cite{mur2015orb} process manually-collected data, active reconstruction methods dynamically plan sensor trajectories to achieve online scene reconstruction while maximizing scene coverage, thus fundamentally requiring real-time generation of high-confidence occupancy maps to distinguish navigable space from obstacles, ensuring collision-free trajectory planning during exploration~\cite{isler2016information,bircher2016receding}.

Existing active reconstruction methods~\cite{feng2024naruto,jin2024gs,xu2024hgs,li2025activesplat,jiang2024fisherrf,jin2025activegs} predominantly rely on depth sensors (e.g., structured-light or Lidars) to directly acquire accurate depth for efficient construction of volumetric occupancy maps like TSDF~\cite{curless1996volumetric} and OctoMap~\cite{hornung2013octomap} for safe spatial reasoning. However, such sensors introduce significant cost, weight, and power constraints that are prohibitive for resource-limited robots/UAVs. 

To advance spatial intelligence, we aim for a vision-only monocular solution. The fundamental challenge is how to acquire dense depth predictions at frame-rate with metric-scale accuracy for safe spatial reasoning. While monocular depth estimation methods~\cite{ranftl2021vision,yang2024depth} lack metric accuracy, recent Multi-View Stereo (MVS) techniques~\cite{sayed2022simplerecon,izquierdo2025mvsanywhere} can estimate high-quality metric depth from well-conditioned image tuples. However, predicting depth with MVS using adjacent frames as a reference is prone to generating sub-optimal results due to insufficient baselines or covisibility. Another issue with MVS is that recent learning-based MVS methods typically process only 8 reference frames, lacking global context, which is a critical limitation since incremental mapping requires globally consistent depth to prevent error accumulation that compromises planning safety and reconstruction quality.

To address these challenges, we introduce ActMVS, the first framework for monocular active scene reconstruction. Inspired by ActiveGS~\cite{jin2025activegs}, we maintain a voxel map for spatial modeling while leveraging Gaussian splatting for high-fidelity reconstruction. In the absence of depth sensors, we develop two key innovations for reliable metric depth estimation: A view factor graph with voxel-frame visibility modeling for informed MVS depth prediction, and a global depth optimization algorithm enforcing cross-view consistency through depth warping and alignment. This enables ActMVS to generate high-quality, globally consistent dense depth maps online, allowing safe trajectory planning and efficient scene reconstruction.\looseness=-1

In summary, our contributions are threefold:
\begin{itemize}
    \item The first work for monocular active reconstruction, with performance competitive with RGB-D methods.
    \item A novel view factor graph formulation with voxel-frame visibility modeling for informed MVS depth prediction.
    \item A novel global depth optimization method leveraging view factor graph to enhance spatial consistency and geometric accuracy.
\end{itemize}

%%%%%%%%%%%%%%%%%%%%%%%%%%%%%%%%%%%%%%%%%%%%%%%%%%%%%%%%%%%%%%%%%%%%%%%%%%%%%%%%%%%%%%%%%%%%%%%%%%%%%%%%%%%%%%%%%%%%%%%%%%%%%%%%%%%%%%%%%%%%%%%%%%%%%%%%%%%%%%%

\section{Related Work}
\label{sec:related}

\subsection{Scene Reconstruction}
Classic 3D scene reconstruction methods~\cite{moulon2016openmvg} such as COLMAP~\cite{schonberger2016structure} utilize multi-view images to reconstruct scenes through complex pipelines involving keypoint detection~\cite{detone2018superpoint}, image matching~\cite{sarlin2020superglue}, camera localization~\cite{brachmann2017dsac}, triangulation~\cite{hartley1997triangulation}, and bundle adjustment~\cite{agarwal2010bundle}, typically representing scenes as sparse point clouds. Classical visual SLAM methods~\cite{zhou2015structslam,mur2017orb,campos2021orb} like ORB-SLAM~\cite{mur2015orb} adopt similar pipelines with greater emphasis on real-time simultaneous localization and mapping.
Recent learning-based approaches have revolutionized scene reconstruction. Neural Radiance Field (NeRF)~\cite{mildenhall2021nerf} and 3DGS (3D Gaussian Splatting)~\cite{kerbl20233d} utilize differentiable scene representations. Feedforward methods~\cite{wang2024dust3r,cabon2025must3r,yang2025fast3r,duisterhof2024mast3r,wang2024vggsfm} such as VGGT~\cite{wang2025vggt} directly predict 3D scenes from images, significantly simplifying reconstruction pipelines. Single-image methods~\cite{pu2023sinmpi, liang2025wonderland, pu2024pano2room, huang2025midi} like CAST~\cite{yao2025cast} achieve scene reconstruction from only a single image, greatly enhancing accessibility. These passive reconstruction methods rely on manually captured data. \looseness=-1

\vspace{-3pt}

\subsection{Active Scene Reconstruction}
Active reconstruction optimizes sensor trajectories to maximize scene coverage during mapping. Traditional methods~\cite{bircher2016receding} focus on enhancing coverage through path planning on voxel maps or meshes, aiming to efficiently navigate through the environment. NeRF-based methods like NARUTO~\cite{feng2024naruto} learn uncertainty grids through hybrid neural representations to guide exploration. 3DGS-based approaches, including GS-Planner~\cite{jin2024gs} and HGS-Planner~\cite{xu2024hgs}, fuse voxel maps with GS-rendered depth estimation. View planning techniques utilize Voronoi diagrams~\cite{li2025activesplat} or Fisher information~\cite{jiang2024fisherrf} to strategically select optimal viewpoints. ActiveGS~\cite{jin2025activegs} represents the state of the art of RGB-D approaches, combining Gaussian splatting with voxel mapping for high-fidelity reconstruction and spatial reasoning.
Critically, all existing active reconstruction methods require depth sensors.

\vspace{-3pt}

\subsection{Depth Estimation}
Monocular depth networks~\cite{ranftl2021vision,yang2024depth} predict scale-ambiguous depth from single images but lack metric accuracy. Multi-View Stereo estimates geometrically consistent depth from posed multi-view images using epipolar geometry~\cite{wang2024learning}. Traditional MVS~\cite{seitz2006comparison} optimizes per-pixel depth hypotheses via photometric consistency, suffering from $O(n^2)$ view-pair complexity. 
Learning-based MVS methods~\cite{yao2018mvsnet,wang2021patchmatchnet,sayed2022simplerecon,izquierdo2025mvsanywhere} enable dense depth prediction and enhance robustness through learnable cost volume aggregation. However, these MVS methods typically process around only 8 reference frames, lacking global context. We adapt the feedforward MVS method MVSA~\cite{izquierdo2025mvsanywhere} for initial dense depth prediction, and further enhance it with a global depth optimization to achieve improved global depth consistency.

%%%%%%%%%%%%%%%%%%%%%%%%%%%%%%%%%%%%%%%%%%%%%%%%%%%%%%%%%%%%%%%%%%%%%%%%%%%%%%%%%%%%%%%%%%%%%%%%%%%%%%%%%%%%%%%%%%%%%%%%%%%%%%%%%%%%%%%%%%%%%%%%%%%%%%%%%%%%%%%
\section{Method}
\subsection{Overview}
\label{ssec:overview}

\begin{figure*}[ht]
    \centering
    \includegraphics[width=0.85\textwidth]{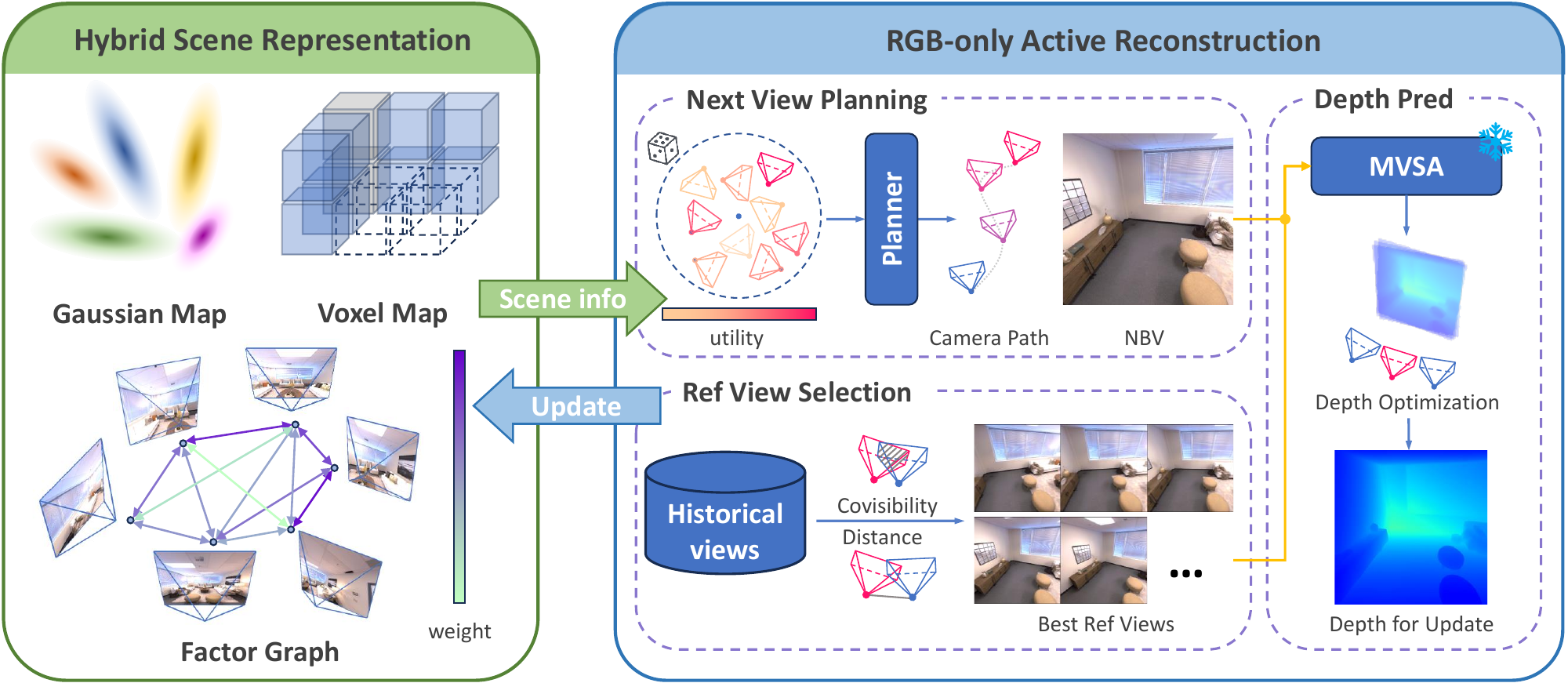}
    % \vspace{-8pt}
    \caption{ActMVS actively reconstructs environments through iterative view planning and incremental mapping. At each timestep $t$, the system accepts posed RGB images ($\mathbf{I}_t$, $\mathbf{P}_t$) as input and outputs a dense depth map $\mathbf{D}_t$ while concurrently updating the View Factor Graph $\mathcal{G}_f$, Voxel Map $\mathcal{M}_v$, and Gaussian Splatting Map $\mathcal{M}_g$.}
    \label{fig:overview}
    \vspace{-10pt}
\end{figure*}

Inspired by ActiveGS~\cite{jin2025activegs}, we leverage a voxel map for spatial representation and a Gaussian-splatting map for high-fidelity surface reconstruction. In the absence of a depth sensor, we introduce a view factor graph incorporating voxel-frame visibility modeling to guide Multi-View Stereo depth prediction during online map updates.

Operating as an online monocular active reconstruction framework, our method sequentially processes image-pose pairs ($\mathbf{I}_t$, $\mathbf{P}_t$) from $t=0$ to $t=T$, where the camera pose $\mathbf{P}_t$ is obtained via simulator measurements during simulation, and via odometry from robots/UAVs during real-world testing, respectively. The algorithm runs until the preset time budget $T$ is reached, and $T$ can be adjusted based on scene size. At each timestep, we generate a reconstructed 3D scene representation comprising the view factor graph $\mathcal{G}_f$, voxel map $\mathcal{M}_v$, and Gaussian map $\mathcal{M}_g$. The overview of our pipeline is shown in Fig.~\ref{fig:overview}.

\subsection{Hybrid Scene Representations}

We employ a View Factor Graph $\mathcal{G}_f = (\mathcal{V}, \mathcal{E})$ to model inter-frame geometric relationships essential for MVS prediction. Nodes $\mathbf{v}_t \in \mathcal{V}$ store view images, optimized depth maps, camera poses and camera intrinsics $\{\mathbf{I}_t, \mathbf{D}_t, \mathbf{P}_t, \mathbf{K}_t\}$, while edges $\mathcal{E}$ encode voxel-frame visibility weights indicating geometric overlap. \looseness=-1

Following~\cite{isler2016information,bircher2016receding}, our Voxel Map $\mathcal{M}_v$ builds upon OctoMap~\cite{hornung2013octomap}. Each voxel stores an occupancy probability $p \in [0,1]$ and updates incrementally using depth-derived point clouds. This representation serves as both an occupancy map for collision-free planning and a visibility model for co-visible surfaces.

We adopt Gaussian surfels~\cite{dai2024high} to construct the Gaussian Splatting Map $\mathcal{M}_g$ for high-fidelity surface modeling. This differentiable representation employs alpha-blending rendering to generate color $\mathbf{I_g}$, depth $\mathbf{D_g}$, normal $\mathbf{N_g}$, and opacity $\mathbf{O_g}$ maps. 

\subsection{Online Active Reconstruction Process}

During initialization, $k$ viewpoints featuring uniform baselines are sampled near the origin to initialize the scene representation. Subsequently, the pipeline iterates between the planning and mapping stages.

\subsubsection{Planning Phase}
We determine the next camera extrinsics $\mathbf{P}$ by selecting the Next-Best-View (NBV) based on a weighted score combining view completeness and travel distance. 

First, candidate viewpoint positions $\mathbf{p}$ are sampled outside the safety radius ($\|\mathbf{p}\| > R_{\min}$), the exploration boundary ($\|\mathbf{p}\| < R_{\max}$), and traversable regions ($\mathcal{M}_v^{\text{(free)}}$).

For each candidate, we calculate the view completeness using the number of visible unexplored voxels, identified by comparing Gaussian-rendered depth maps against the projections of voxel centroids. The travel distance $d_{\text{travel}}$ to the candidate is computed via an A* path search~\cite{hart1968formal} over the voxel map $\mathcal{M}_v$.

Finally, the candidate viewpoint with the highest weighted score is chosen as the NBV. Instead of navigating directly to the NBV, we interpolate collision-free trajectories along the computed A* path. This ensures sufficient covisibility for subsequent MVS depth prediction.

\subsubsection{Mapping Phase}
First, visibility modeling computes pairwise overlaps $\Omega_{tj}$ via voxel covisibility. The graph $\mathcal{G}_f$ updates by inserting $\{\mathbf{I}_t, \mathbf{P}_t\}$ and connecting to the top-$k$ neighbors with edge weights $\Omega_{tj}$. 

Subsequently, we adapt MVSA~\cite{izquierdo2025mvsanywhere} to estimate dense depth $\widehat{\mathbf{D}}_t$ with selected neighbors as reference views, followed by $N_{\text{iter}}$ steps of global depth optimization optimizes over $\mathbf{D}_t$ through pairwise depth warping in $\mathcal{G}_f$. 

Finally, we update $\mathcal{M}_v$ via probabilistic occupancy integration. For $\mathcal{M}_g$, we perform Gaussian primitive initialization via depth point-cloud projection in low-opacity areas where rendered opacity satisfies $\mathbf{O_g} < 0.1$, following $\mathcal{M}_g$ optimization by minimizing photometric and depth $\mathcal{L}_1$ losses against all $\{\mathbf{I}_t, \mathbf{D}_t\}$ in $\mathcal{G}_f$.

%%%%%%%%%%%%%%%%%%%%%%%%%%%%%%%%%%%%%%%%%%%%%%%%%%%%%%%%%%%%%%%%%%%%%%%%%%%%%%%%%%%%%%%%%%%%%%%%%%%%%%%%%%%%%%%%%%%%%%%%%%%%%%%%%%%%%%%%%%%%%%%%%%%%%%%%%%%%%%%
% ----------------------- hanyx 7.16

\subsection{Voxel-Frame Visibility Modeling}
\label{subsec:visibility}
\begin{figure}[t]
    \centering
    \includegraphics[width=0.45\textwidth]{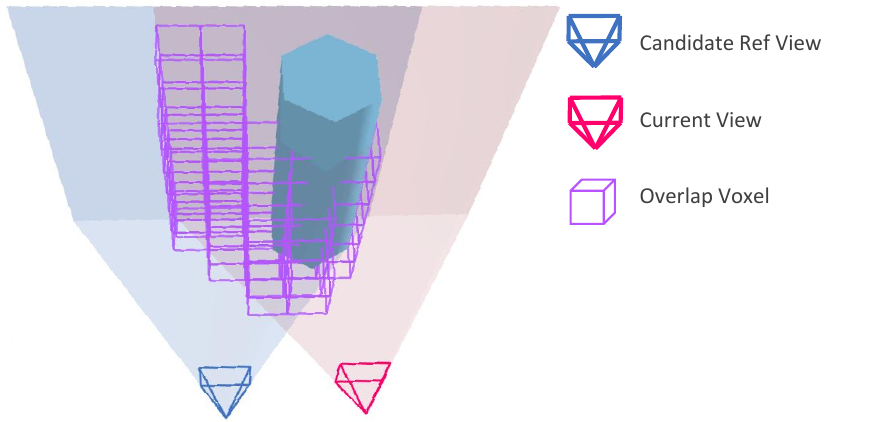}
    % \vspace{-10pt}
    \caption{Visualization of voxels in the overlapped region of the current view and candidate reference view. Obstructed voxels  are excluded to prevent interference caused by the area invisible to either of the two viewpoints.}
    \vspace{-15pt}
    \label{fig:frustum}
\end{figure}

MVSA predicts 3D-consistent depth by effectively incorporating multi-view geometry constraint and monocular feature from multiple posed images. Therefore, evaluating the covisibility between viewpoints during active reconstruction process is crucial for selecting reference frames that provide effective information, ultimately achieving more accurate depth prediction. A naive solution is to rely purely on similarity in camera position and orientation. However, this method does not account for occlusions. Even viewpoints with similar camera positions and orientations may fail to provide sufficient RGB overlap due to obstructions nearby. Another approach involves feature extraction and matching to find the suitable viewpoints directly in the feature space. Nevertheless, this method lacks robustness in regions with similar textures and introduces significant computational overhead. Based on the geometric priors contained in the gaussian map and the efficient scene representation provided by our voxel map, we propose a frame visibility assessment method that considers occlusions while ensuring high computational efficiency, as shown in Fig.~\ref{fig:frustum}. 

To be specific, we compute per-frame frustum masks $M_{t}$ that identify voxels visible from each viewpoint in 3 steps. (1) Render gaussian depth using current camera extrinsic and intrinsic matrix  \((P_{t}, K_{t})\) to obtain depth map \(D_{t}\). (2) Project voxels onto the camera plane, obtaining UV coordinates \((u_{k}, v_{k})\) and depth values \(d_{k}\) of each voxel center \(\vartheta_{k}\). (3) Select voxels on the camera plane with depth values that lie within corresponding rendered gaussian depth. The whole process can be formulated as:

\begin{equation}
    (u_{k}, v_{k}, d_{k}) = \psi_{P_{t}, K_{t}}(\vartheta_{k} ), \vartheta_{k} \in \mathcal{M}_{v},
\end{equation}
\vspace{-15pt}
\begin{equation}
    M_{t} = \{k|d_{k} \in (0, D_{t}[u_{k}, v_{k}]) \wedge u_{k} \in [0, w) \wedge v_{k} \in [0, h) \},
\end{equation}
where \(\psi_{K_{t}, P_{t}}\) denotes the rendering process using camera extrinsic and intrinsic \((P_{t}, K_{t})\).

This establishes the fundamental voxel-to-frame relationship that underpins our spatial reasoning and provides efficient overlap calculation by a single Logical AND operation.

% \begin{equation}
%     M_i = f_{\text{frustum}}(\mathbf{T}_i, \mathbf{K}_i, D_i, \mathcal{V})
% \end{equation}

% where $\mathbf{T}_i$ is the camera pose (extrinsic matrix), $\mathbf{K}_i$ is the intrinsic matrix, $D_i$ is the depth map, and $\mathcal{V}$ is the voxel grid. The visibility computation involves:

% This establishes the fundamental voxel-to-frame relationship that underpins our spatial reasoning.

% -----------------------------

\textit{Inter-Frame Relation Modeling:} Without ground truth depth, ensuring geometrically consistent depth predictions across multiple views is particularly crucial for accurate geometry reconstruction. MVSA incorporates the camera pose information of the reference frame into the cost volume, providing the model with implicit cues about scale derived from camera translation and rotation. To enhance the 3D consistency of depth predictions, we calculate covisibility between source and reference frames as:
\begin{equation}
    \Omega_{ij} = |M_i \cap M_j| = \sum_{k=1}^{|\mathcal{M}_v|} \mathbb{I}[M_i^k = 1 \land M_j^k = 1],
\end{equation}
where $\mathbb{I}$ is the indicator function. % This measures shared map coverage between frames.

Then, we quantify the correlation score between frame i and frame j using the following formula balancing both covisibility and scale information contained in camera translation:
\begin{equation}
\left\{
\begin{aligned}
    \lambda_{1} \Omega_{ij} + \lambda_{2}\|\mathbf{t}_i - \mathbf{t}_j\|_2, & \quad \Omega_{ij} \geq \epsilon \\
    0, & \quad else
\end{aligned}
,
\right.
\end{equation}
where \(\epsilon\) is a threshold set to 50, discarding viewpoints with no frustum overlap. 

% The pairwise relationship between frame $i$ and $j$ is quantified using:

% \begin{equation}
%     r_{ij} = \phi(\text{spatial\_dist}(i,j), \text{angular\_dev}(i,j), \text{voxel\_overlap}(i,j))
% \end{equation}

% [?] need subsection or not
% \subsubsection{Spatial-Angular Constraints}
% We enforce dual constraints when selecting reference frames:

Although MVSA performs camera metadata normalization to maintain scene scale-agnostic properties, our experiments reveal that selecting reference frames with poses too close to the input frame can magnify discrepancies in depth predictions. Conversely, reference frames with excessively distant camera poses introduce features with poor similarity and degrade depth accuracy. To address this, we impose a spatial-angular constraint on camera translation $ \delta_{\min} \leq \|\mathbf{t}_i - \mathbf{t}_j\|_2 \leq \delta_{\max} $ and orientation $ \theta_{\min} \leq \arccos(\mathbf{n}_i \cdot \mathbf{n}_j) \leq \theta_{\max} $,
where $\mathbf{t}$ denotes the camera position and $\mathbf{n}$ is the optical axis direction. This ensures that the frames have meaningful baselines for triangulation.

% \subsubsection{Voxel Covisibility Metric}
% The primary relational metric is the voxel overlap coefficient:

% \begin{equation}
%     \Omega_{ij} = |M_i \cap M_j| = \sum_{k=1}^{|\mathcal{V}|} \mathbb{I}[M_i^k = 1 \land M_j^k = 1]
% \end{equation}

% where $\mathbb{I}$ is the indicator function. This measures shared map coverage between frames.

% TODO: describe frame choosing num

% -----------------------------

%%%%%%%%%%%%%%%%%%%%%%%%%%%%%%%%%%%%%%%%%%%%%%%%%%%%%%%%%%%%%%%%%%%%%%%%%%%%%%%%%%%%%%%%%%%%%%%%%%%%%%%%%%%%%%%%%%%%%%%%%%%%%%%%%%%%%%%%%%%%%%%%%%%%%%%%%%%%%%%

\subsection{Globally Optimized Depth}
\label{subsec:global_depth_opt}

\begin{figure}[t]
    \centering
    \includegraphics[width=0.45\textwidth]{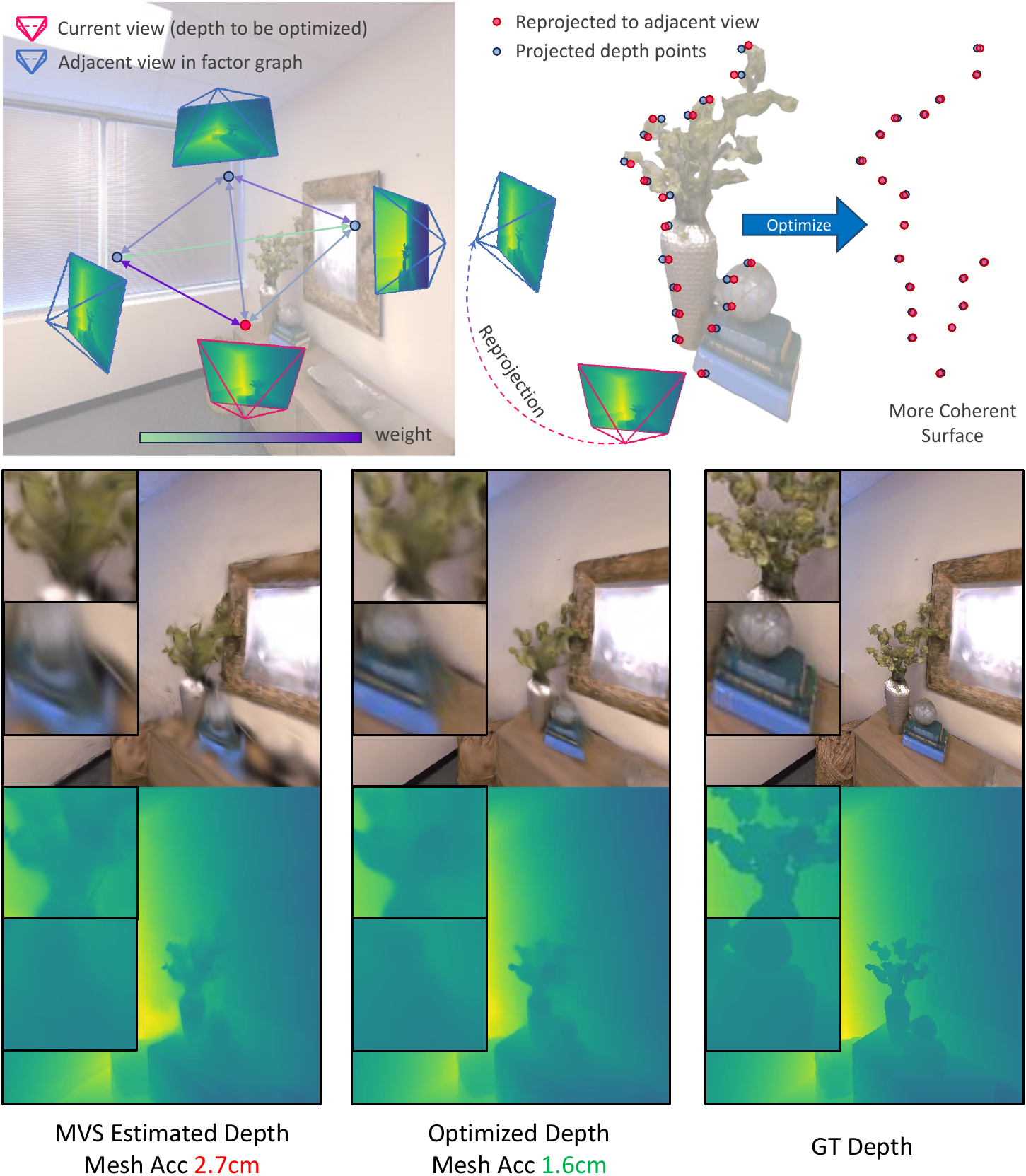}
    % \vspace{-10pt}
    \caption{Our global depth optimization over view factor graph enforces 3D consistency in co-visible regions through depth warping alignment, resulting in enhanced mesh quality.}
    \label{fig:Opt}
    \vspace{-15pt}
\end{figure}

Even with our searched well-conditioned reference frames, MVSA typically processes only 8 reference frames for prediction, lacking global context, which is prone to error accumulation that compromises planning safety and reconstruction quality. To address this issue, we propose a depth optimization algorithm based on our view factor graph $\mathcal{G}_f$ to achieve globally consistent depth estimates. We optimize initial depth predictions by applying pixel-level multi-view constraints to adjacent frames within $\mathcal{G}_f$. This approach enforces 3D consistency in co-visible regions via depth warping alignment, as illustrated in Fig.~\ref{fig:Opt}.

\subsubsection{Depth Consistency Factor}

The core constraint $\mathcal{F}_\text{depth}$ enforces geometric consistency between frame pairs $(i,j) \in \mathcal{E}$ of the view factor graph $\mathcal{G}_f$. For a pair $(i,j)$ with corresponding nodes $\mathbf{v}_i=\{\mathbf{I}_i, \mathbf{D}_i, \mathbf{P}_i, \mathbf{K}_i\}$ and $\mathbf{v}_j=\{\mathbf{I}_j, \mathbf{D}_j, \mathbf{P}_j, \mathbf{K}_j\}$, we aim to align their co-visible 3D structures. This is achieved by backprojecting depth estimates to 3D, warping points between views, and imposing alignment constraints on valid correspondences. Specifically, the depth consistency factor is computed as follows. For a pixel $\mathbf{p}_i = (u_i,v_i)$ in $\mathbf{v}_i$, its 3D position in the camera coordinate system is:
\begin{equation}
\mathbf{X}_i^\text{c}(\mathbf{p}_i) = \mathbf{K}_i^{-1} 
\begin{bmatrix}
u_i \cdot z_i \\
v_i \cdot z_i \\
z_i
\end{bmatrix}, \quad \text{where} \quad z_i = \mathbf{D}_i(\mathbf{p}_i)
\end{equation}
with $\mathbf{K}_i$ denoting the camera intrinsic matrix. This point transforms to the frame $j$'s camera coordinate system by:
\begin{equation}
\mathbf{X}_j^\text{c} = \mathbf{P}_j \mathbf{P}_i^{-1} \mathbf{X}_i^\text{c}
\end{equation}
where $\mathbf{P}$ denotes world-to-camera transformation matrices.

The projective correspondence $\mathbf{p}_j = (u_j,v_j)$ in frame $j$ is given by perspective projection:
\begin{equation}
\mathbf{p}_j = \pi\left( \mathbf{K}_j \mathbf{X}_j^\text{c} \right) = \left( 
\frac{ \mathbf{K}_{j(1,:)} \mathbf{X}_j^\text{c} }{ \mathbf{K}_{j(3,:)} \mathbf{X}_j^\text{c} },
\frac{ \mathbf{K}_{j(2,:)} \mathbf{X}_j^\text{c} }{ \mathbf{K}_{j(3,:)} \mathbf{X}_j^\text{c} }
\right),
\end{equation}
where $\mathbf{K}_{j(k,:)}$ denotes the $k$-th row of $\mathbf{K}_j$.

For robustness against outliers, we formulate the depth consistency factor using a log-space Huber loss:
\begin{equation}
\mathcal{F}_{\text{depth}}^{i,j} = \sum_{\mathbf{p}_i \in \mathbf{X}^\text{c}_i} \rho_\delta\left( \log \mathbf{D}_j(\mathbf{p}_j) - \log z_j^\prime \right) \cdot \mathbf{M}_{i,j}(\mathbf{p}_i)
\end{equation}
where $z_j^\prime = \mathbf{X}_j^\text{c}(z)$ is the transformed depth value. $\rho_\delta$ denotes the Huber norm ($\delta$ specifies the transition point). $\mathbf{M}_{i,j}$ is a binary validity mask excluding occlusions $z_j^\prime > \mathbf{D}_j(\mathbf{p}_j) + \tau_o$, boundary violations $\mathbf{p}_j \notin [1, W-1] \times [1, H-1]$ and invalid depths $\mathbf{D}_j(\mathbf{p}_j) \leq \tau_d \lor z_j^\prime \leq \tau_d$, where $\tau_o$ and $\tau_d$ are predefined tolerance thresholds.

\subsubsection{Regularization Factor}

As depth warping yields sparse pixel-level depth map supervision, we incorporate total variation regularization to ensure smoothness and geometric plausibility:
\begin{equation}
\mathcal{F}_\text{reg-TV}^i = \lambda_\text{TV} \sum_{\mathbf{p}_i} \left( \|\nabla_u \mathbf{D}_i\|_\gamma + \|\nabla_v \mathbf{D}_i\|_\gamma \right),
\end{equation}
where $\|\cdot\|_\gamma$ denotes the Charbonnier penalty $\|\cdot\|_\gamma = \sqrt{(x)^2 + \gamma^2}$
, $\nabla_u$ and $\nabla_v$ represent horizontal/vertical gradient operators. $\gamma$ controls edge sensitivity.

\subsubsection{Optimization}

The complete optimization minimizes:
\begin{equation}
\underset{\{\mathbf{D}_i\}_{i\in\mathcal{K}}}{\min} \left[ \sum_{(i,j) \in \mathcal{E}} \mathcal{F}_{\text{depth}}^{i,j} + \sum_{i \in \mathcal{K}} \mathcal{F}_{\text{reg-TV}}^i \right],
\end{equation}
where $\mathcal{K}$ denotes keyframes selected from $\mathcal{V}$ based on spatial distribution and visual covisibility.

We solve this non-linear optimization via iterative backpropagation using the Adam optimizer. Sequential per-frame optimization across $\mathcal{G}_f$ ensures global consistency through incremental geometric constraint propagation while maintaining computational efficiency.

%%%%%%%%%%%%%%%%%%%%%%%%%%%%%%%%%%%%%%%%%%%%%%%%%%%%%%%%%%%%%%%%%%%%%%%%%%%%%%%%%%%%%%%%%%%%%%%%%%%%%%%%%%%%%%%%%%%%%%%%%%%%%%%%%%%%%%%%%%%%%%%%%%%%%%%%%%%%%%%

\section{Experiments}

\begin{figure}[t]
    \centering
    \includegraphics[width=0.45\textwidth]{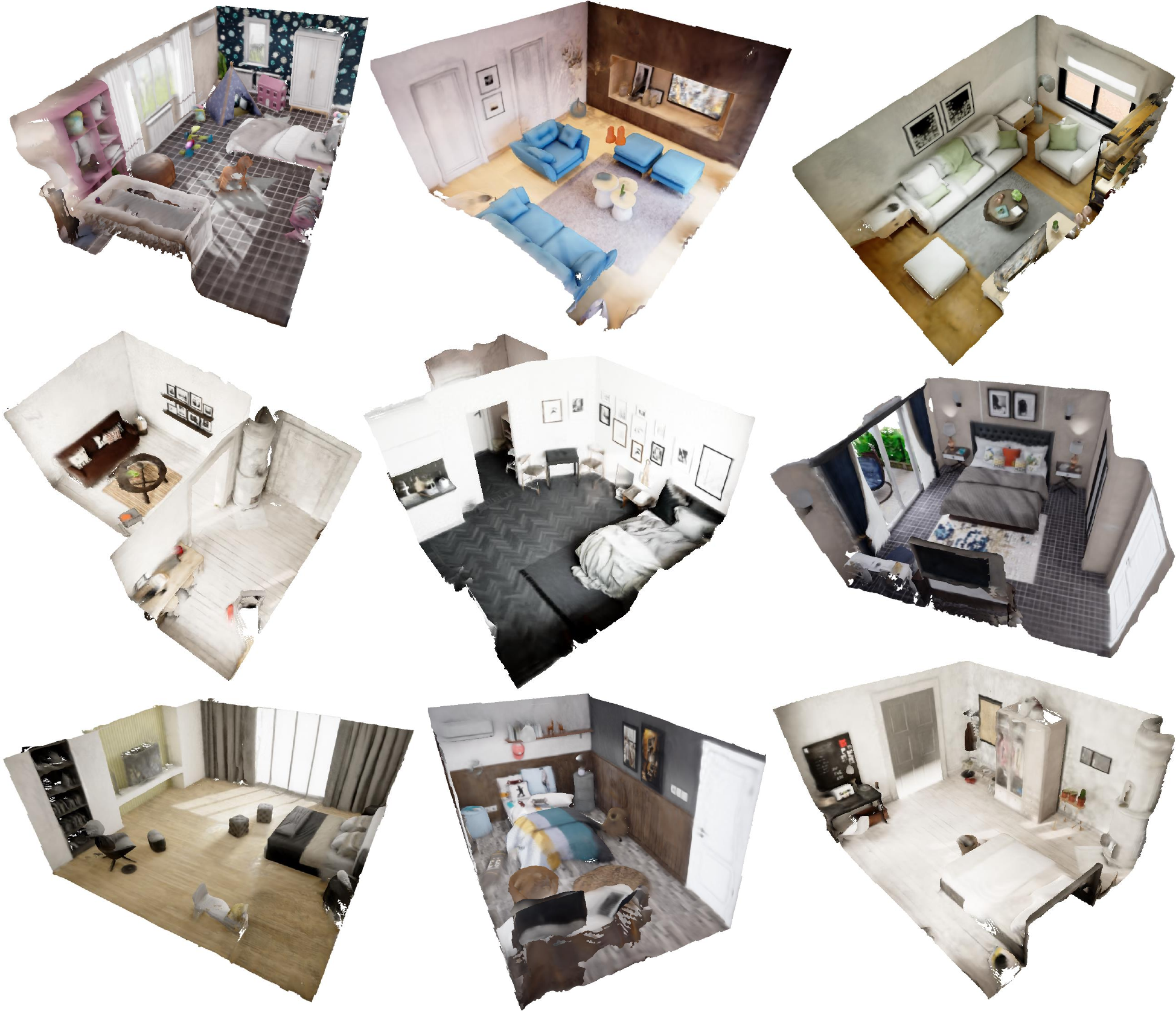}
    % \vspace{-10pt}
    \caption{Reconstruction results of UAV deployment in Airsim simulation.}
    \label{fig:airsim}
    \vspace{-15pt}
\end{figure}

\begin{table}[t]
  \centering
  \scriptsize
  \setlength{\tabcolsep}{2pt}
  \renewcommand{\arraystretch}{0.8}
  \caption{Quantitative evaluation on Replica scenes. Methods are grouped by input modality (RGBD vs. RGB).}
  \label{tab:combined_results}
  \begin{adjustbox}{max width=0.45\textwidth}
  \begin{tabular}{lcccc}
    \toprule
    \multirow{2}{*}{\textbf{Method}} & 
    \multicolumn{4}{c}{\textbf{Scene}} \\
    \cmidrule{2-5}
    & 
    \multicolumn{2}{c}{\textbf{Room0}} & 
    \multicolumn{2}{c}{\textbf{Room1}} \\
    \cmidrule(lr){2-3} \cmidrule(l){4-5}
    & \textbf{Render} & \textbf{Mesh} & \textbf{Render} & \textbf{Mesh} \\
    \midrule

    \textbf{RGBD} & & & & \\
    ~~ActiveGS 
    & \textbf{28.14} \textbf{0.868} \textbf{0.199} & \textbf{1.081} \textbf{1.278} \textbf{1.180} 
    & \textbf{29.55} \textbf{0.881} \textbf{0.178} & \textbf{0.902} \textbf{1.063} \textbf{0.983} \\
    ~~Naruto 
    & 25.51 0.769 0.380 & 1.907 1.619 1.763 
    & 27.23 0.792 0.372 & 1.642 1.296 1.469 \\
    \midrule

    \textbf{RGB} & & & & \\
    ~~ActMVS 
    & 26.32 0.839 \textbf{0.263} & \textbf{2.211} \textbf{2.583} \textbf{2.397} 
    & \textbf{28.05} \textbf{0.857} \textbf{0.242} & \textbf{1.654} 2.207 \textbf{1.930} \\
    ~~w/o OPT 
    & \textbf{27.17} \textbf{0.843} 0.266 & 5.066 2.608 3.837 
    & 26.24 0.830 0.258 & 2.705 \textbf{2.081} 2.393 \\
    ~~w/o REF
    & 23.35 0.792 0.314 & 4.387 4.758 4.573 
    & 24.06 0.792 0.329 & 32.12 4.421 18.271 \\
    ~~w/o MVS 
    & 11.47 0.521 0.723 & 43.66 48.79 46.225 
    & 12.64 0.561 0.702 & 38.04 41.21 39.625 \\
    \midrule

    & 
    \multicolumn{2}{c}{\textbf{Room2}} & 
    \multicolumn{2}{c}{\textbf{Office0}} \\
    \cmidrule(lr){2-3} \cmidrule(l){4-5}
    & \textbf{Render} & \textbf{Mesh} & \textbf{Render} & \textbf{Mesh} \\
    \midrule

    \textbf{RGBD} & & & & \\
    ~~ActiveGS 
    & \textbf{30.01} \textbf{0.906} \textbf{0.154} & \textbf{0.940} \textbf{0.970} \textbf{0.955} 
    & \textbf{34.62} \textbf{0.950} \textbf{0.089} & \textbf{0.896} \textbf{1.004} \textbf{0.950} \\
    ~~Naruto 
    & 27.50 0.827 0.322 & 1.553 1.543 1.548 
    & 30.55 0.877 0.304 & 1.489 1.350 1.420 \\
    \midrule

    \textbf{RGB} & & & & \\
    ~~ActMVS 
    & 27.33 0.875 0.217 & \textbf{1.854} \textbf{1.901} \textbf{1.878} 
    & \textbf{31.63} \textbf{0.927} \textbf{0.148} & \textbf{2.111} 2.343 \textbf{2.227} \\
    ~~w/o OPT 
    & 26.90 0.864 0.230 & 2.359 2.267 2.313 
    & 31.62 0.925 0.168 & 8.411 \textbf{2.000} 5.206 \\
    ~~w/o REF
    & \textbf{28.69} \textbf{0.891} \textbf{0.204} & 2.086 2.116 2.101 
    & 28.01 0.890 0.224 & 5.036 3.914 4.475 \\
    ~~w/o MVS 
    & 10.60 0.465 0.703 & 45.98 52.31 49.145 
    & 15.15 0.511 0.631 & 40.17 43.48 41.825 \\
    \midrule

    & 
    \multicolumn{2}{c}{\textbf{Office1}} & 
    \multicolumn{2}{c}{\textbf{Office2}} \\
    \cmidrule(lr){2-3} \cmidrule(l){4-5}
    & \textbf{Render} & \textbf{Mesh} & \textbf{Render} & \textbf{Mesh} \\
    \midrule

    \textbf{RGBD} & & & & \\
    ~~ActiveGS 
    & \textbf{34.15} \textbf{0.949} \textbf{0.104} & \textbf{0.752} \textbf{0.876} \textbf{0.814} 
    & \textbf{30.10} \textbf{0.915} \textbf{0.135} & \textbf{1.059} \textbf{1.143} \textbf{1.101} \\
    ~~Naruto 
    & 30.21 0.885 0.260 & 1.188 1.082 1.135 
    & 25.17 0.807 0.334 & 2.618 2.409 2.514 \\
    \midrule

    \textbf{RGB} & & & & \\
    ~~ActMVS 
    & \textbf{31.96} \textbf{0.921} \textbf{0.147} & \textbf{2.227} 4.590 \textbf{3.409}
    & \textbf{26.82} \textbf{0.884} \textbf{0.197} & \textbf{3.246} \textbf{3.347} \textbf{3.297} \\
    ~~w/o OPT 
    & 27.67 0.873 0.216 & 35.97 \textbf{4.352} 20.161 
    & 22.54 0.798 0.308 & 6.787 4.213 5.500 \\
    ~~w/o REF
    & 28.93 0.894 0.190 & 5.106 5.948 5.527 
    & 18.94 0.777 0.336 & 7.909 8.745 8.327 \\
    ~~w/o MVS 
    & 16.46 0.423 0.498 & 38.38 54.56 46.47 
    & 10.10 0.526 0.675 & 41.61 39.77 40.69 \\
    \midrule

    & 
    \multicolumn{2}{c}{\textbf{Office3}} & 
    \multicolumn{2}{c}{\textbf{Office4}} \\
    \cmidrule(lr){2-3} \cmidrule(l){4-5}
    & \textbf{Render} & \textbf{Mesh} & \textbf{Render} & \textbf{Mesh} \\
    \midrule

    \textbf{RGBD} & & & & \\
    ~~ActiveGS 
    & \textbf{29.91} \textbf{0.910} \textbf{0.146} & \textbf{1.201} \textbf{1.578} \textbf{1.390} 
    & \textbf{31.75} \textbf{0.918} \textbf{0.145} & \textbf{1.166} \textbf{1.337} \textbf{1.252} \\
    ~~Naruto 
    & 24.93 0.793 0.369 & 2.706 2.255 2.481 
    & 27.60 0.844 0.330 & 2.148 1.747 1.948 \\
    \midrule

    \textbf{RGB} & & & & \\
    ~~ActMVS 
    & \textbf{26.18} \textbf{0.869} \textbf{0.222} & \textbf{2.880} 3.751 \textbf{3.316} 
    & 29.75 0.896 0.201 & 2.382 3.189 2.786 \\
    ~~w/o OPT 
    & 24.06 0.833 0.290 & 38.31 \textbf{3.258} 20.784 
    & \textbf{29.99} \textbf{0.903} \textbf{0.199} & \textbf{2.247} \textbf{2.336} \textbf{2.292} \\
    ~~w/o REF
    & 22.69 0.824 0.310 & 10.66 5.612 8.136 
    & 28.87 0.889 0.208 & 3.293 4.105 3.699 \\
    ~~w/o MVS 
    & 11.00 0.529 0.675 & 38.90 45.51 42.205 
    & 13.06 0.589 0.612 & 42.94 41.21 42.075 \\
    \bottomrule
  \end{tabular}
  \end{adjustbox}
\end{table}

\begin{figure*}[ht]
    \centering
    \includegraphics[width=0.95\textwidth]{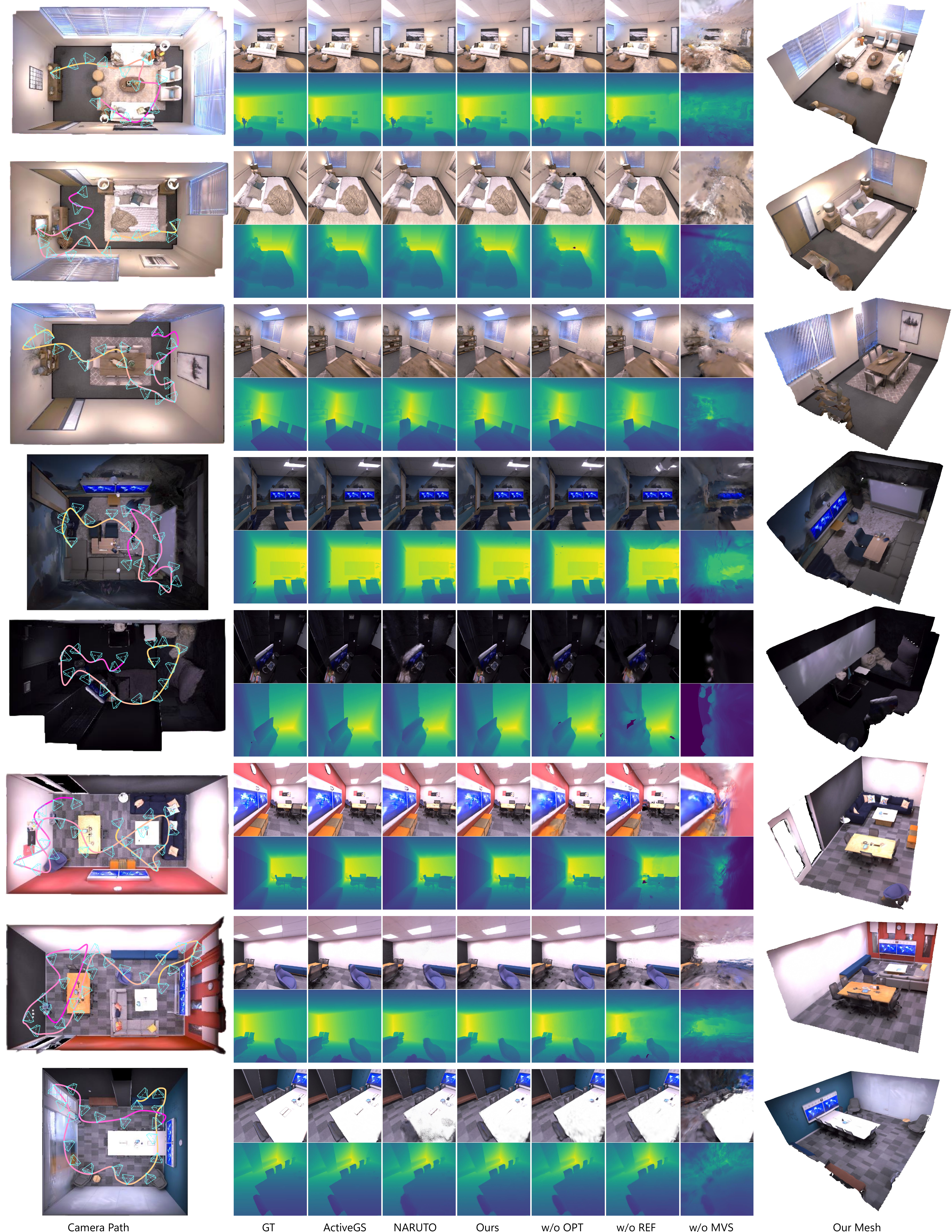}
    % \vspace{-0pt}
    \caption{Qualitative comparison on Replica scenes showcase the camera path and the reconstructed mesh of ours, along with ground truth RGB-D and rendered RGB-D of ActiveGS, NARUTO, our full method, and ablation variants from randomly sampled novel viewpoints. While ActiveGS exhibits the closest visual resemblance to ground truth, our method produces closely high-quality outputs. For better visualization of the 3D scene in all figures in this paper, mesh faces blocking the room interior are deleted. Please zoom in for detailed inspections.}
    % \vspace{-10pt}
    \label{fig:comparison}
\end{figure*}

\subsection{Evaluation}

\subsubsection{Implementation Details}
The occupancy grid $\mathcal{M}_v$ employs $20$~cm\textsuperscript{3} voxels.
Exploration radius bounds are $R_{\min} = 0.3$~m and $R_{\max} = 0.5$~m. 
The NBV score weights view completeness ($w_{\text{comp}} = 1.0$) against travel distance ($w_{\text{dist}} = 0.5$).
Depth estimation employs MVSA~\cite{izquierdo2025mvsanywhere} at $512 \times 512$ resolution.
Spatial-angular constraints enforce baseline thresholds $\delta_{\min} = 0.1$~m, $\delta_{\max} = 0.5$~m, and angular bounds $\theta_{\min} = 0^\circ$, $\theta_{\max} = 25^\circ$. Correlation parameters $\lambda_1 = 1.0$ and $\lambda_2 = 0.1$ prioritize overlap percentage over  translation, with voxel overlap threshold $\epsilon = 50$ preventing degenerate pairs. Each input frame selects the top-16 reference frames by correlation scores.
Each keyframe undergoes $N_{\text{iter}} = 20$ global depth optimization steps with Huber norm $\delta = 0.1$, Charbonnier penalty $\gamma = 0.001$, TV regularization $\lambda_\text{TV} = 0.1$, occlusion detection threshold $\tau_o = 0.1$~m, and depth validity threshold $\tau_d = 0.01$~m.
Gaussian map $\mathcal{M}_g$ undergoes 10 optimization iterations per step with batch size 8 over randomly selected views. 
We conduct evaluation experiments of all methods on an NVIDIA L40 48GB GPU with Intel Xeon Platinum 8370C 2.80GHz CPU.

\subsubsection{Baselines and Protocols}
Due to the absence of prior monocular methods for active reconstruction, we evaluate against two state-of-the-art RGB-D baselines ActiveGS and NARUTO. Experiments utilize the Habitat simulator~\cite{savva2019habitat} and Replica dataset~\cite{straub2019replica} with $90^\circ$ field of view and $512 \times 512$ resolution. 
Rendering quality is evaluated using standard metrics PSNR $\uparrow$, SSIM $\uparrow$, LPIPS $\downarrow$. Mesh reconstruction performance is assessed via Accuracy $\downarrow$ (cm), Completion $\downarrow$ (cm), and Chamfer distance $\downarrow$ (cm).

\subsubsection{Quantitative Evaluation}
Quantitative results across eight Replica scenes in Table~\ref{tab:combined_results} categorize methods by depth-sensor dependency (RGB vs. RGB-D). The RGB group consists of our method and relevant ablation studies. Rendering results are computed from 1000 novel viewpoints randomly sampled per scene. On average, our approach achieves the best performance within the RGB group across both rendering and mesh quality metrics while closely approaching RGB-D baselines. These results validate that our multi-view constrained depth optimization effectively compensates for the absence of depth sensors.

\subsubsection{Qualitative Evaluation}
Visual comparisons in Fig.~\ref{fig:comparison} showcase the camera path and the reconstructed mesh of ours, along with ground truth RGB-D and rendered RGB-D of ActiveGS, NARUTO, our full method, and ablation variants from randomly sampled novel viewpoints. While ActiveGS exhibits the closest visual resemblance to ground truth, our method produces closely high-quality outputs. NARUTO’s NeRF-based renderings suffer significant degradation in novel views, due to the sparse viewpoints inherent in fast UAV motion trajectories.

\subsubsection{Discussions} 
ActiveGS achieves the fastest convergence with an average runtime of 300 seconds. Due to NeRF's slower performance, NARUTO requires 1800 seconds on average. Compared to ActiveGS, our method necessitates denser camera trajectories and additional computation for view factor graph construction, depth prediction and optimization, resulting in an average runtime of 1200 seconds.
We recommend checking our supplementary video for detailed camera paths, as well as depth prediction results and Gaussian update process during the reconstruction.

\subsection{Ablation Studies}
Ablation studies are quantified in Table~\ref{tab:combined_results} and visualized in Fig.~\ref{fig:comparison}, validating critical component contributions.
\subsubsection{w/o OPT} Removing global depth optimization causes depth instability and surface degradation. This significantly reduces mesh quality, demonstrating that global optimization enforces consistency across viewpoints and stabilizes reconstructed geometry.
\subsubsection{w/o REF} Replacing our voxel-based reference frame search with simple adjacent-frame selection reduces the quality and stability of MVS depth predictions. The suboptimal baseline lengths between frames often cause MVS failures due to insufficient parallax for reliable scale recovery.
\subsubsection{w/o MVS} We replace the MVS prediction in our method with the state-of-the-art monocular metric depth estimator DepthAnythingV2~\cite{yang2024depth}. Sole reliance on DepthAnythingV2 metric depth estimation model causes catastrophic reconstruction failures due to significant error accumulation. This validates that integrating Multi-View Stereo is essential for achieving plausible geometry under strong perspective changes.

% %%%%%%%%%%% hanyx 07.21
\subsection{Extended Simulations}
Real-world applicability is demonstrated through AirSim simulations featuring quadrotor UAV navigation in more realistic and complex environments compared to Habitat. However, the large amount of video memory required by 3DGS-based reconstruction exceeds the capacity of current onboard devices. To approximate real-world deployment, we run ActMVS on a server equipped with a single NVIDIA L40 GPU and use Flask for the network transmission of waypoint and camera image data. Instead of relying on controller APIs provided by Airsim, we opted for the widely used open-source flight control system PX4 with Robot Operating System (ROS) and MAVLink for sensor data processing and transmission, simulating a more authentic operating environment for real-world drones. In Airsim simulation, we utilize Ego-planner~\cite{zhou2020ego} to execute the camera path generated by ActMVS, ensuring compliance with the dynamical constraints of a real drone. Reconstruction results in Fig.~\ref{fig:airsim} demonstrate the effectiveness of ActMVS in various scenarios, where the drone did not encounter any collisions during the experiment, further validating the feasibility of the ActMVS path planning.

% ↑ HERE IS A SIMPLIFIED VERSION ↑
% Real-world applicability is demonstrated through AirSim simulations featuring quadrotor UAV navigation in photorealistic indoor environments. Due to video memory limitation of current onboard device, we run ActMVS on a server equipped with a single NVIDIA L40 GPU and use Flask for the network transmission of waypoint and camera image data. The open-source flight control system PX4 with Robot Operating System (ROS) and MAVLink were used for sensor data processing and transmission, which simulates a more authentic operating environment for real-world drones. In Airsim simulation, we utilize Ego-planner to execute the camera path generated by ActMVS, ensuring compliance with the dynamical constraints of a real drone. Reconstruction results (Fig.~\ref{fig:airsim}) demonstrate the performance of ActMVS in various scenarios, where the drone did not encounter any collisions during the experiment, further validating the feasibility of the ActMVS path planning.

%%%%%%%%%%

\section{Conclusion}
We present the first monocular scene reconstruction framework that enables robust autonomous aerial reconstruction without depth sensors, by integrating a view factor graph with voxel-frame visibility modeling and a global depth optimization mechanism that leverages the connectivity within this graph to enforce cross-view consistency. Experiment results demonstrate competitive performance against RGB-D baselines, closely approaching them in both rendering fidelity and geometric accuracy. Future work will focus on real-time deployment optimization for fully autonomous drone operations under physical constraints.

%%%%%%%%%%%%%%%%%%%%%%%%%%%%%%%%%%%%%%%%%%%%%%%%%%%%%%%%%%%%%%%%%%%%%%%%%%%%%%%%

\bibliographystyle{IEEEtran}

\bibliography{aaai2026}

\end{document}